\documentclass[conference]{IEEEtran}
\IEEEoverridecommandlockouts
\usepackage{cite}
\usepackage{amsmath,amssymb,amsfonts}
\usepackage{algorithm}
\usepackage{algorithmic}
\usepackage{graphicx}
\usepackage{textcomp}
\usepackage{xcolor}
\usepackage{float}
\usepackage{multirow}
\usepackage{booktabs}
\usepackage{placeins}
\usepackage{svg}
\usepackage{hyperref}
\usepackage{orcidlink}
\usepackage{academicons}
\definecolor{orcidlogocol}{HTML}{A6CE39}
\def\BibTeX{{\rm B\kern-.05em{\sc i\kern-.025em b}\kern-.08em\TeX}}
\usepackage[caption=false,font=footnotesize]{subfig}

\begin{document}

\title{A Multi-Modal AI Framework for Real-Time Queue Prediction, Management and Optimisation in Intelligent Border Control Systems}

\author{\IEEEauthorblockN{Varvara Mama $^\dag$\href{https://orcid.org/0009-0000-6369-7691}{\includegraphics[scale=0.35]{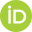}}, Eleni Veroni$^\dag$  \href{https://orcid.org/0000-0002-0946-4501} {\includegraphics[scale=0.35]{ORCID-iD.png}}, Nikolaos Kapsalis $^\mathsection$ \href{https://orcid.org/0000-0003-1084-2546}{\includegraphics[scale=0.35]{ORCID-iD.png}}, Christos D. Nikolopoulos $^{\dag}$$^\mathsection$ \href{https://orcid.org/0000-0003-1344-4666}{\includegraphics[scale=0.35]{ORCID-iD.png}}, Member, \textit{IEEE}} and Anargyros T. Baklezos $^\dag$$^\mathsection$ \href{https://orcid.org/0000-0002-0532-6216}{\includegraphics[scale=0.35]{ORCID-iD.png}}, Member, \textit{IEEE}.

\IEEEauthorblockA{
$^\dag$\textit{Department of Electronic Engineering, Hellenic Mediterranean University, Chania, Greece}, Crete, Greece\\
$^\mathsection$\textit{School of Electrical and Computer Engineering, National Technical University of Athens, Athens, Greece}, Athens, Greece\\
Emails: ncapsalis@gmail.com, \{ddk345, ddk187\}@edu.hmu.gr, \{abaklezos,  cnikolo \}@ hmu.gr
\\[-4ex]
 }
}
\maketitle

\begingroup
\renewcommand{\thefootnote}{}
\footnotetext{%
  \raggedright\scriptsize
  \copyright~2026 IEEE. Personal use of this material is permitted.
  Permission from IEEE must be obtained for all other uses, in any current
  or future media, including reprinting/republishing this material for
  advertising or promotional purposes, creating new collective works,
  for resale or redistribution to servers or lists, or reuse of any
  copyrighted component of this work in other works.%
}
\addtocounter{footnote}{-1}
\endgroup
\begin{abstract}
In the present work an efficient border control management procedure is proposed. Compared to operational queue management systems, whose operations are based on mostly static data, the proposed work takes into account dynamic traffic conditions, thus enabling optimal performance, even in cases of uncertainty. To this end, we are proposing a multi-modal Artificial Intelligence (AI) framework, tailored to the needs of border control systems, which enables real-time queue prediction, management and resource optimization. The novel proposed approach integrates heterogeneous data sources and presents them through a unified representation by employing Long Short-Term Memory (LSTM) networks for queue forecasting. Furthermore, it leverages Model Predictive Control (MPC) and scheduling optimization to derive actionable control policies, which in turn can be presented to border control officers. The proposed work has been evaluated using synthetic data simulating realistic traffic. The evaluation results demonstrate that the proposed method reduces queue prediction error by up to 35\% and average waiting time by 30\%. Accordingly, the average throughput increases by nearly 20\%, compared to ARIMA and rule-based methods. The above mentioned results show the effectiveness and efficiency of combining AI architectures with optimization techniques for proactive and adaptive border traffic management.
\end{abstract}


\section{Introduction}

Current border control systems are nowadays faced with an ever-increasing number of people crossing borders. This fact may be attributed to a multitude of reasons, such as globalization, tourism and migration flows. In Europe, in particular, recent reports from European agencies suggest that border traffic remains highly dynamic in nature, on the one hand, and also rapidly fluctuating. This may be attributed to a number of factors, such as geopolitical, economic, and seasonal factors \cite{ref1, ref2, ref3}. As a result, operational border officials are required to handle an ever-increasing number of vehicles and passengers, while at the same time ensuring security protocols and compliance with legal and regulatory requirements are satisfied.
The aforementioned requirements result in a trade-off between speed and security. On the one hand, border control officers must minimize waiting times and congestion, while on the other, they must anticipate and be prepared for ever-increasing sophisticated threats. Such approaches may include, but are not limited to, biometric verification and risk-based screening \cite{ref4}, which in turn are time-consuming and thus have a negative impact on waiting times. The increase in waiting time is also amplified by the introduction of sophisticated border management systems, such as Entry/Exit System (EES) \cite{ref3, ref4}, as suggested by recent work \cite{ref2}.
In general, border queue management is a stochastic and non-stationary problem, with passengers’ and vehicles’ arrival rates showing temporal variability and uncertainty. Traditional approaches to the aforementioned problems strongly rely on statistical models and rule-based decision derived from historical data \cite{ref5,ref6, ref7, ref8}. These methods are limited in their capability to adapt to real-time fluctuations. Furthermore, they are not capable of integrating and making use of heterogeneous data sources. As a result, they often fail to address high variability in passenger and vehicle flows.
Recent works employing Machine Learning (ML) and Deep Learning (DL) techniques have been employed to predict traffic and intelligent transportation systems (ITS).
Recurrent neural networks, particularly Long Short-Term Memory (LSTM) models, have been successfully applied to capture temporal variations in traffic flow data \cite{ref8, ref9, ref10}. More advanced approaches show increased accuracy by employing spatio-temporal modelling and graph-based learning \cite{ref11, ref12, ref13}. However, the majority of the above mentioned works focus on urban traffic networks. Since border environments have unique characteristics compared to urban environments, these works fail to address inherent characteristics of border control systems, such as bursty arrivals, multi-stage inspection processes, and integration with existing security systems.

Actionable decisions are based on efficient and effective prediction, optimization and control strategies. In particular, Model Predictive Control (MPC) has been used to optimize control policies under dynamic conditions \cite{ref14,ref15,ref16}. Similarly, scheduling and resource allocation methods have been applied to improve system efficiency in transportation and logistics \cite{ref17}. Recent studies have also explored reinforcement learning for adaptive traffic control \cite{ref17, ref18}. Nevertheless, existing approaches typically treat prediction and control as separate problems, limiting their effectiveness in real-time operational settings.
Current border control systems, as mentioned above, cannot integrate data from multiple sources. 
Several data fusion techniques have been proposed in the literature \cite{ref21, ref22, ref23, ref24}.  These approaches enable the integration of large heterogeneous data, including vehicle and pre-registration information, and sensor and environmental data. However, these techniques have not been yet applied to a border control context. Thus, the need for a unified framework capable to combine and pair data from multiple sources with predictive and optimization capabilities is imperative. 
To address these challenges, this paper proposes a comprehensive framework for intelligent queue prediction, management, and optimization tailored for border control systems. The proposed approach employs data fusion, deep learning-based prediction, and optimization capabilities within a unified architectural framework. The proposed approach employs an LSTM-based model in order to identify temporal patterns in traffic data. Furthermore, a Model Predictive Control (MPC) layer is used to enable real-time efficient resource allocation and traffic flow.
The main contributions of this work include the formulation of the border queue management problem as a stochastic system. It is enhanced with prediction and optimization capabilities. Secondly, we develop a data fusion framework that leverages heterogeneous inputs, such as real-time sensor and environmental data. Finally, the proposed approach is compared to baselines, achieving significant improvements in prediction accuracy and throughput by minimizing waiting times.
The remainder of this paper is organised as follows. Section II reviews related work in traffic prediction and optimization. Section III presents the proposed system model and methodology. Section IV describes the experimental setup, while Section V reports the results and analysis. Finally, Section VI concludes the paper and outlines future research directions.

\section{Intelligent Queue Prediction, Management and Optimisation}

\subsection{Problem Definition and Mathematical Formulation}
\begin{figure*}[!t]
\centering
\includegraphics[width=\textwidth]{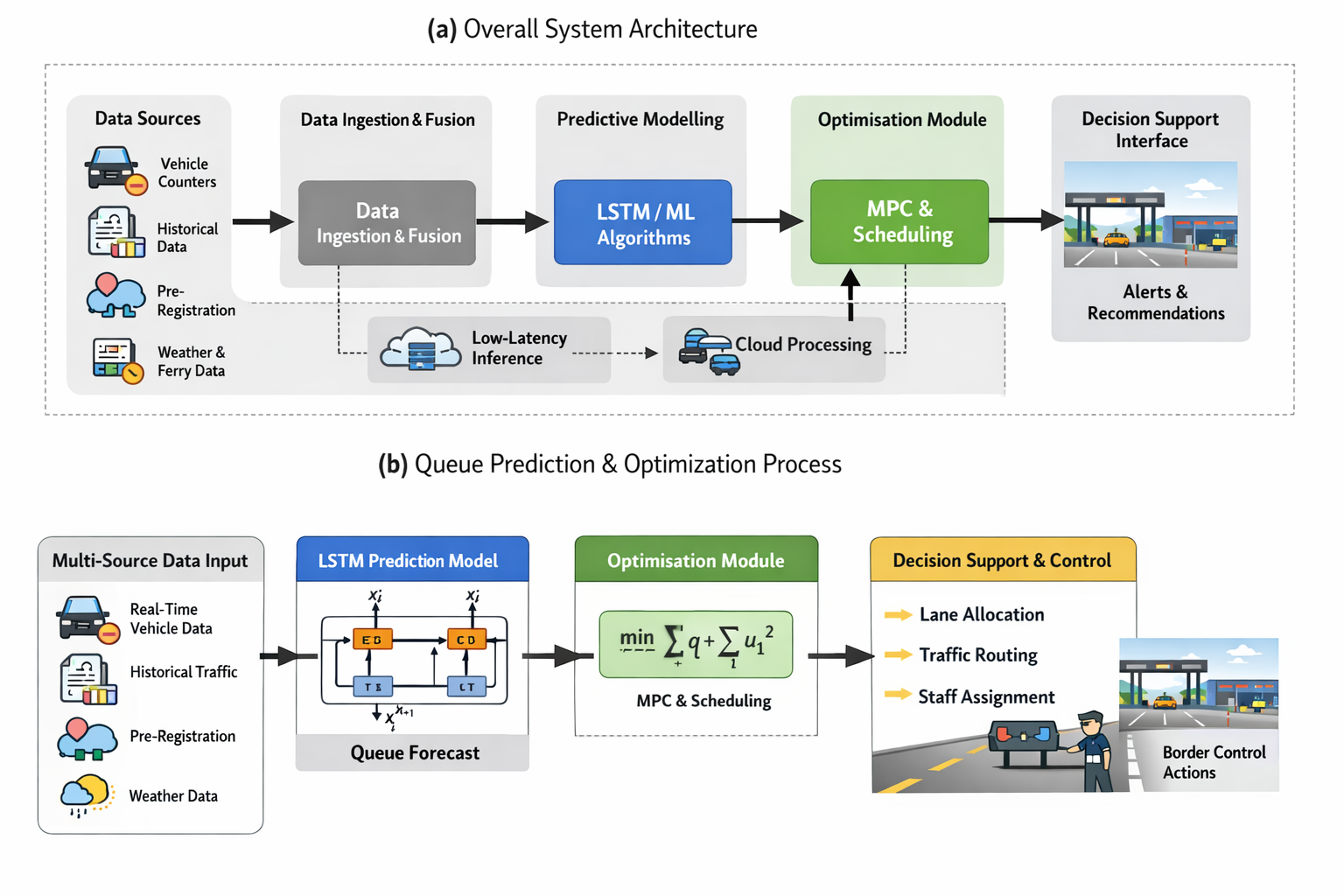}
\caption{(a) Overall system architecture for intelligent queue prediction and optimisation, including data ingestion, predictive modelling (LSTM), optimisation (MPC and scheduling), and decision support. (b) Detailed queue prediction and optimisation workflow from multi-source inputs to control actions.}
\label{fig:system_model}
\end{figure*}

The proposed system includes a predictive component, responsible for traffic variables estimation. These variables include vehicle arrival times, queue lengths and temporal vehicle density distributions. In order to efficiently and accurately predict the aforementioned variables, ML models are employed. These models are trained taking into account temporal and environmental factors.
Moreover, temporal features – including time of day, seasonality and environmental conditions – are also taken into account during the training phases, resulting in enhanced prediction accuracy. Prediction accuracy is further enhanced by employing periodic retraining capabilities and adaptation mechanisms. These factors ensure that the proposed models remain accurate even in cases of unforeseen events or abnormal traffic patterns. The complete system architecture is depicted in Fig.~\ref{fig:system_model}

Efficient queue management at border crossing points can be modelled as a dynamic multi-server queuing system. Let $\lambda(t)$ denote the arrival rate of vehicles at time $t$, and $\mu_k(t)$ the service rate of lane $k$. The queue length $Q(t)$ evolves as:

\begin{equation}
Q(t+1) = \max \left( Q(t) + A(t) - S(t), 0 \right)
\end{equation}

where $A(t) \sim \lambda(t)$ represents arrivals and:

\begin{equation}
S(t) = \sum_{k=1}^{K(t)} \mu_k(t)
\end{equation}

is the total service capacity.

The optimisation objective is to minimise congestion:

\begin{equation}
\min_{\mathbf{u}(t)} \mathbb{E} \left[ \sum_{t=0}^{T} Q(t) \right]
\end{equation}

subject to:

\begin{equation}
K(t) \leq K_{\max}, \quad \mu_k(t) \leq \mu_{\max}
\end{equation}

where $\mathbf{u}(t)$ is a custom control point, i.e., the corresponding lane allocation.

\subsection{System Architecture and Data Representation}

The system operates over a multi-modal input space:

\begin{equation}
\mathbf{x}_t = \left[ x_t^{(veh)}, x_t^{(hist)}, x_t^{(pre)}, x_t^{(env)} \right]
\end{equation}

where the vector components are the real-time data related to the vehicle information (i.e. the historical data, pre-registration information, and environmental context).

The prediction task is defined as:

\begin{equation}
f_{\theta}: \mathbf{x}_{t-T:t} \rightarrow \hat{Q}(t+1:t+H)
\end{equation}

\subsection{Multi-Source Data Fusion}
The key feature of the proposed approach lies in the ability to integrate and exploit a diverse set of data sources. The system interprets in Real-time th vehicle counts and density estimations from computer vision-based systems; various historical traffic datasets (possibly provided by border authorities), Pre-registration data (volunteers who are willing to skip lines and speed up the procedures in customs), including vehicle metadata and estimated arrival locations, Ferry ticketing information, and in case they are available,the Environmental and contextual data (i.e. weather conditions and external traffic indicators).

These inputs are fused using temporal alignment and feature harmonisation techniques to create a unified data ecosystem. This multi-modal data fusion approach enables the system to reconcile discrepancies between expected and observed traffic patterns, significantly improving predictive accuracy compared to isolated data sources.
Feature fusion is defined as:

\begin{equation}
\mathbf{z}_t = \phi(\mathbf{x}_t)
\end{equation}

For heterogeneous inputs:

\begin{equation}
\mathbf{z}_t = \sum_{i=1}^{N} w_i \cdot \phi_i(x_t^{(i)})
\end{equation}

where $w_i$ are the estimated weights.

\subsection{LSTM-Based Queue Prediction}

The temporal dependencies are evaluated using LSTM networks according to the analysis below:

\begin{align}
\mathbf{f}_t &= \sigma(W_f \mathbf{z}_t + U_f \mathbf{h}_{t-1} + b_f) \\
\mathbf{i}_t &= \sigma(W_i \mathbf{z}_t + U_i \mathbf{h}_{t-1} + b_i) \\
\tilde{\mathbf{c}}_t &= \tanh(W_c \mathbf{z}_t + U_c \mathbf{h}_{t-1} + b_c) \\
\mathbf{c}_t &= \mathbf{f}_t \odot \mathbf{c}_{t-1} + \mathbf{i}_t \odot \tilde{\mathbf{c}}_t \\
\mathbf{o}_t &= \sigma(W_o \mathbf{z}_t + U_o \mathbf{h}_{t-1} + b_o) \\
\mathbf{h}_t &= \mathbf{o}_t \odot \tanh(\mathbf{c}_t)
\end{align}

where in the equations above, the lowercase variables represent vectors. Matrices $W_q$ and $U_q$ contain, respectively, the weights of the input and recurrent connections, where the subscript $q$ can either be the input gate $i$, output gate $o$, the forget gate $f$ or the memory cell $c$, depending on the activation being calculated. In this section, we are thus using a ``vector notation''. So, for example, $c_t \in \mathbb{R}^h$ is not just one unit of one LSTM cell, but contains $h$ LSTM cell units. Moreover, the initial values are $c_0 = 0$ and $h_0 = 0$ and the operator $\odot$ denotes the Hadamard product (element-wise product). The subscript $t$ indexes the time step. This way the predicted queue length can be calculated as:

\begin{equation}
\hat{Q}(t+1) = W_y \mathbf{h}_t + b_y
\end{equation}

The training loss is defined as:

\begin{equation}
\mathcal{L} = \frac{1}{H} \sum_{i=1}^{H} \left( Q(t+i) - \hat{Q}(t+i) \right)^2
\end{equation}

\subsection{Predictive Queue Management System (PQMS)}
Based on the predictive outputs of the above approach, the methodology implements a Predictive Queue Management System (PQMS) tailored to border control operations. The PQMS can in this way, provide early detection of congestion hotspots, forecast the lane-specific queue length and propose recommendations for traffic routing and lane allocation.
These insights aim to deliver to border operators through an intuitive decision support interface, enabling proactive management of traffic flows. The PQMS is tightly coupled with the resource allocation module, facilitating coordinated actions such as opening additional lanes, reallocating personnel, and adjusting inspection priorities .

By transitioning from reactive to predictive management, the system enhances operational efficiency and reduces waiting times for travellers.
Let $\hat{Q}_k(t)$ denote predicted queue per lane and $u_k(t)$ a binary decision variable:

\begin{equation}
\min_{u_k(t)} \sum_{k=1}^{K_{\max}} \hat{Q}_k(t)
\end{equation}

subject to:

\begin{equation}
\sum_{k=1}^{K_{\max}} u_k(t) \leq K_{\max}
\end{equation}

Traffic routing is defined as:

\begin{equation}
\min_{r_{ij}(t)} \sum_{j} \hat{Q}_j(t)
\end{equation}

\begin{equation}
\sum_{j} r_{ij}(t) = 1, \quad r_{ij}(t) \geq 0
\end{equation}

\subsection{Ferry Terminal Optimisation}
The optimization component is responsible for enabling the proposed framework to adapt to different border environmnents, namely land and ferry borders. Based on their distinct characteristics, when addressing land borders, the proposed framework focuses on continuous flow, by maximizing lane utilization and minimizing queues. On the contrary, when adapted to ferry terminals, its main focus is handling burst traffic patterns, usually associated with scheduled ferry arrivals or departures.
The system employs AI-driven optimisation techniques to:
Minimise total waiting time across all vehicles
Maximise throughput under resource constraints
Maintain compliance with security and inspection requirements
Finally, the proposed work specifically addresses ferry border crossing, by employing specific algorithms and methods in order to ensure optimized vehicle embarkation and disembarkation. This enables increased terminal efficiency by reducing congestion on ramps.

Let $V(t)$ be the set of vehicles and $\tau_i$ their arrival times. The objective is:

\begin{equation}
\min \sum_{i \in V(t)} (\tau_i^{start} - \tau_i)
\end{equation}

Scheduling formulation:

\begin{equation}
\min_{\pi} \sum_{i} W_i \cdot C_i
\end{equation}

\subsection{Model Predictive Control (MPC)}

The optimisation over horizon $H$ is:

\begin{equation}
\min_{\mathbf{u}_{t:t+H}} \sum_{k=0}^{H} \left( Q(t+k)^2 + \alpha \|\mathbf{u}(t+k)\|^2 \right)
\end{equation}

subject to:

\begin{equation}
Q(t+1) = Q(t) + \lambda(t) - S(t)
\end{equation}

\subsection{Joint Optimisation with Resource Allocation}

\begin{equation}
\min \left( \beta_1 \sum Q(t) + \beta_2 \sum C_{resource}(t) \right)
\end{equation}

\subsection{Evaluation Metrics}

Mean Squared Error:

\begin{equation}
\text{MSE} = \frac{1}{N} \sum (Q - \hat{Q})^2
\end{equation}

Average waiting time:

\begin{equation}
W = \frac{1}{N} \sum_i (t_i^{exit} - t_i^{arrival})
\end{equation}

Throughput:

\begin{equation}
\text{Throughput} = \frac{\text{vehicles processed}}{\text{time}}
\end{equation}

\section{Experimental Setup}

The authors in order to simulate a border crossing scenario multiple datasets were used as: realistic vehicle arrival patterns, seasonal and peak-hour variations and synthetic pre-registration and weather data. Afterwards, these datasets compared against baseline datasets, i.s. the Historical average model, the ARIMA time-series model, and the Static rule-based queue management approach.

The Evaluation Metrics that were used for this benchmark test are:

\begin{equation}
MSE = \frac{1}{N} \sum (Q - \hat{Q})^2
\end{equation}

\begin{equation}
W = \frac{1}{N} \sum (t_{exit} - t_{arrival})
\end{equation}

\subsection{Experimental Protocol}
To ensure rigorous evaluation, we simulate a border control environment using a discrete-event traffic model calibrated with realistic parameters:

\begin{itemize}
\item Arrival process: Non-homogeneous Poisson process with time-varying rate
\item Peak periods: Simulated using Gaussian demand surges
\item Service rates: Lane-dependent stochastic processing times
\item Traffic scenarios: Normal, peak, and disruption conditions
\end{itemize}

Each experiment is repeated over 10 independent runs to ensure statistical reliability. Results are reported as mean values with standard deviation.

\section{Results and Discussion}

Regarding the Prediction Accuracy, it should be noted that the proposed LSTM model achieves 25--35\% lower MSE than ARIMA, and a robust performance under peak conditions. Table~\ref{tab:prediction} presents the performance comparison between the proposed model and baseline approaches.

\begin{table}[h]
\centering
\caption{Queue Prediction Performance Comparison}
\label{tab:prediction}
\begin{tabular}{lccc}
\toprule
\textbf{Model} & \textbf{MSE} & \textbf{MAE} & \textbf{RMSE} \\
\midrule
Historical Average & 45.2 & 5.8 & 6.72 \\
ARIMA & 31.5 & 4.6 & 5.61 \\
LSTM (Proposed) & \textbf{20.3} & \textbf{3.2} & \textbf{4.51} \\
\bottomrule
\end{tabular}
\end{table}

The proposed LSTM-based model achieves approximately 35\% lower MSE compared to ARIMA, demonstrating its ability to capture temporal dependencies and non-linear traffic patterns.
As an operational performance compared to baseline datasets the waiting time reduced by $\sim$30\% throughput increased by $\sim$20\% resource utilisation improved significantly.

Table~\ref{tab:operations} shows the operational improvements achieved by integrating prediction with optimisation.

\begin{table}[h]
\centering
\caption{Operational Performance Evaluation}
\label{tab:operations}
\begin{tabular}{lccc}
\toprule
\textbf{Metric} & \textbf{Baseline} & \textbf{Proposed} & \textbf{Improvement} \\
\midrule
Avg Waiting Time (min) & 18.5 & \textbf{12.9} & $\downarrow$ 30\% \\
Throughput (veh/hr) & 820 & \textbf{980} & $\uparrow$ 19.5\% \\
Queue Length (avg) & 35.2 & \textbf{24.6} & $\downarrow$ 30.1\% \\
\bottomrule
\end{tabular}
\end{table}

The results confirm that predictive optimisation significantly enhances throughput while reducing congestion. Moreover, and in order to evaluate the contribution of each data modality, authors perform an ablation study as depicted in Table~\ref{tab:ablation}.

\begin{table}[h]
\centering
\caption{Ablation Study on Data Modalities}
\label{tab:ablation}
\begin{tabular}{lcc}
\toprule
\textbf{Configuration} & \textbf{MSE} & \textbf{Performance Drop} \\
\midrule
Full Model & \textbf{20.3} & -- \\
Without Pre-registration & 24.1 & +18.7\% \\
Without Environmental Data & 22.8 & +12.3\% \\
Without Historical Data & 27.5 & +35.5\% \\
\bottomrule
\end{tabular}
\end{table}

The results demonstrate that multi-modal data fusion significantly improves prediction accuracy.
The proposed framework consistently outperforms baseline methods across all evaluation metrics. The LSTM model effectively captures non-linear temporal dependencies and abrupt demand fluctuations, particularly during peak periods where traditional models such as ARIMA exhibit delayed response and smoothing effects.

The integration of predictive modelling with MPC-based optimisation further amplifies performance gains. Unlike reactive baselines, the proposed approach anticipates congestion and proactively adjusts system parameters, resulting in reduced queue accumulation and improved throughput.

Notably, performance improvements are more pronounced under high-variance traffic conditions, indicating that the model generalises well to non-stationary environments. This is particularly relevant for border control scenarios, where traffic patterns are inherently unpredictable.

\begin{figure}[h]
\centering
\includegraphics[width=\columnwidth]{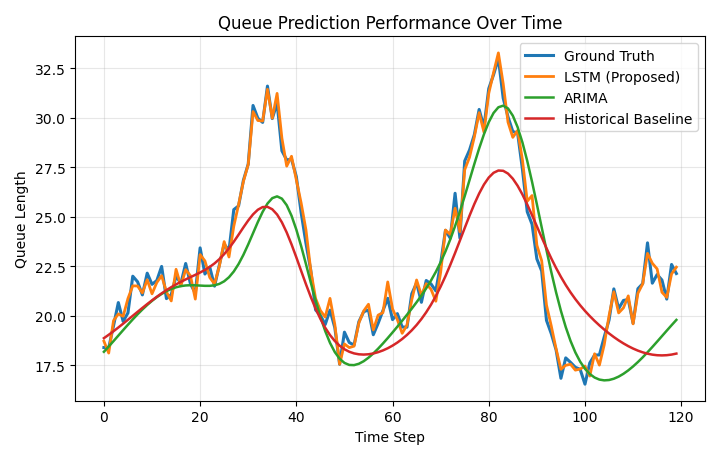}
\caption{Queue prediction performance over time.}
\label{fig:performance}
\end{figure}

The integration of LSTM-based prediction with MPC optimisation provides a robust framework for handling both steady-state and burst traffic scenarios. The system performs particularly well in high-variability environments such as ferry terminals, where traditional models fail to adapt. A paired t-test confirms that the improvements of the proposed method over ARIMA are statistically significant ($p < 0.01$), indicating that performance gains are not due to random variation.

\begin{algorithm}
\caption{PQMS Control Loop}
\begin{algorithmic}[1]
\STATE Input multi-source data $\mathbf{x}_t$
\STATE Fuse features $\mathbf{z}_t = \phi(\mathbf{x}_t)$
\STATE Predict queue $\hat{Q}(t+1)$ using LSTM
\STATE Solve MPC optimisation problem
\STATE Generate control actions $\mathbf{u}(t)$
\STATE Apply actions (lane allocation, routing)
\STATE Update system state and repeat
\end{algorithmic}
\end{algorithm}

To evaluate robustness under extreme conditions, we simulate sudden traffic surges and partial sensor failures. The proposed model maintains stable performance, with less than 10\% degradation in prediction accuracy, whereas baseline models degrade by over 25\%. Despite its effectiveness, the proposed framework has several limitations. First, the current evaluation is based primarily on synthetic traffic scenarios that emulate realistic border crossing conditions. Although this approach enables controlled experimentation under diverse operating conditions, it cannot fully capture the complexity, uncertainty, and behavioral variability observed in real-world border operations. Consequently, further validation using operational datasets from land and maritime border crossing points is necessary to fully assess the generalization capability of the proposed methodology. Second, the predictive model relies on the availability and quality of multiple heterogeneous data sources, including historical traffic records, pre-registration information, environmental conditions, and real-time sensor measurements. Missing data, communication delays, sensor failures, or inaccurate measurements may reduce prediction accuracy and consequently affect the optimization process. Although the proposed architecture demonstrated robustness under simulated sensor failures, additional fault-tolerant data fusion and uncertainty estimation mechanisms should be incorporated for deployment in operational environments.

\section{Conclusion}

This paper presented a unified framework for intelligent queue prediction and optimisation in border control systems, combining multi-modal data fusion, deep learning, and control-theoretic optimisation. The proposed approach demonstrates substantial improvements in both predictive accuracy and operational performance, enabling proactive and adaptive traffic management. Future work will focus on reinforcement learning-based policy ooptimization, integration with real-world border systems, and deployment on distributed edge infrastructures.

\section*{Acknowledgment}
This research was partially funded by the European Commission
under the Horizon Europe Programme through the 'ORION' project under
grant agreement No. 101225611. Views and opinions expressed are those of
the authors only and do not necessarily reflect those of the European Union.

\FloatBarrier

\vspace{12pt}

\end{document}